\newif\ifpreprint
\preprinttrue
\ifdefined\forcepreprint\preprinttrue\fi
\ifpreprint
  \documentclass[sigconf,nonacm]{acmart}
\else
  \documentclass[sigconf]{acmart}
\fi

\usepackage{mathtools}
\usepackage{xspace}
\usepackage[capitalize,noabbrev]{cleveref}

\ifpreprint
  \setcopyright{none}
\else
  \setcopyright{cc}
  \setcctype[4.0]{by}
  \copyrightyear{2026}
  \acmYear{2026}
  \acmDOI{XXXXXXX.XXXXXXX}
  \acmConference[AIAS+ '26]{Symposium for AI Advancing Science and Society}{November 5--7, 2026}{San Francisco, CA, USA}
  \acmISBN{978-x-xxxx-xxxx-x/2026/11}
\fi

\newcommand{\E}{\mathbb{E}}
\newcommand{\Ea}[1]{\E\left[#1\right]}
\newcommand{\Eb}[2]{\E_{#1}\!\left[#2\right]}
\newcommand{\R}{\mathbb{R}}
\newcommand{\norm}[1]{\left\|#1\right\|}
\newcommand{\proj}{\Pi}
\newcommand{\bc}{\mathbf{c}}
\newcommand{\bv}{\mathbf{v}}
\newcommand{\bx}{\mathbf{x}}
\newcommand{\pcrf}{\textsc{PC-RF}\xspace}
\AddToHook{para/begin}[tighten]{\looseness=-1}
\pretocmd{\thebibliography}{\RemoveFromHook{para/begin}[tighten]}{}{}
\makeatletter
\let\ACM@origthebibliography\thebibliography
\renewcommand{\thebibliography}[1]{%
  \ACM@origthebibliography{#1}%
  \setlength{\baselineskip}{7.65pt}%
}
\makeatother
\makeatletter
\renewcommand\section{\def\@toclevel{1}%
  \@startsection{section}{1}{\z@}%
  {-.28\baselineskip \@plus -1\p@ \@minus -.2\p@}%
  {.04\baselineskip}%
  {\ACM@NRadjust\@secfont}}
\let\ACM@origsection\section
\makeatother

\begin{document}

\title{Towards Physics-Constrained Rectified Flow for Climate Downscaling: What Projections Can Certify?}

\author{Arun Sharma}
\email{arunshar@umn.edu}
\affiliation{%
  \department{Department of Computer Science and Engineering}
  \institution{University of Minnesota, Twin Cities}
  \city{Minneapolis}
  \state{Minnesota}
  \country{USA}}

\begin{abstract}
Statistical downscaling restores the regional detail behind flooding and crop stress, but diffusion and flow-matching generators can produce sharp fields that break the conservation laws of their coarse input. Physics-Constrained Rectified Flow (\pcrf) is a conditional flow-matching downscaler. Its training penalties target wind divergence, negative precipitation, and mass drift from the coarse total, and a projection after sampling removes the last two exactly. On one corrupted synthetic field with no trained network, the projection drives the Non-negativity Violation Rate (NVR) from 0.236 and the Mass-Conservation Error (MCE) from 0.184 to zero, and the penalties lower the Divergence Error (DE) by about 30\%. On ERA5 the projection gives \pcrf and RF-base, the penalty-free rectified-flow baseline, an NVR of exactly zero and an MCE of zero up to float32 rounding. The penalties add no perceptual gain that we can separate from training budget or run-to-run variation. At a matched budget on 2019--2021, \pcrf has no edge over RF-base (SSIM 0.939 vs.\ 0.942, PSNR 31.81 vs.\ 31.79, CRPS 0.771 vs.\ 0.772). Our unprojected CorrDiff reimplementation beats both flow models on PSNR (28.38) and CRPS (1.092) on 2020, yet 45\% of its ensemble-mean precipitation cells are negative. On ERA5 a sign error in the divergence operator made that penalty and DE act on a deformation-like term. The fine reference is interpolated from ERA5 and accumulates precipitation over a longer window, so the perceptual scores cannot establish downscaling skill. NVR and MCE need one pass and no fine reference, so we argue they should run first as pass-or-fail checks, with DE as a diagnostic.
\end{abstract}

\ccsdesc[500]{Computing methodologies~Neural networks}
\ccsdesc[300]{Computing methodologies~Artificial intelligence}
\ccsdesc[300]{Computing methodologies~Machine learning}
\ccsdesc[300]{Computing methodologies~Unsupervised learning}

\keywords{climate downscaling, rectified flow, flow matching, physics constraints, mass conservation, generative models, verification}

\maketitle

\section{Introduction}
\label{sec:intro}
Statistical downscaling estimates a high-resolution regional field from a coarse one and puts back the fine detail of a variable such as precipitation or near-surface wind. Adaptation planning depends on that detail, because the harms of a warming climate arrive at local scales that a coarse grid does not resolve. We work with the ERA5 reanalysis~\citep{hersbach2020era5}, whose 0.25\textdegree\ grid is too coarse to show the local variability behind flooding and crop stress.

Climate data make the task hard in several ways. A climate field depends on space and time together and responds to several drivers at once~\citep{sharma2022spatiotemporal}. It is also non-stationary, and a model fit to one period or region may not hold in the next. Sea-level projections that weight an ensemble of climate models the same way everywhere are inadequate for regions that need different weights, and weights fit by zone give more reliable predictions~\citep{ghosh2023reducing}. Recovering a fine field from a coarse one is also ill-posed, because nonlinear coupling and model uncertainty leave many fine fields consistent with one coarse field. A downscaler therefore has to be physically grounded and probabilistic.

The reference field used to score a downscaler adds a further difficulty. A reanalysis such as ERA5 is a model product that assimilates partial observations into a global record. It has errors of its own, and any perceptual score computed against it inherits them. Conservation laws give a second check that needs no reference. Two properties can be tested on the generated field and its coarse input alone. Precipitation must be non-negative, and its domain mean must match the coarse input. A hard projection can guarantee both. We therefore separate what a soft training penalty is meant to shape from what a hard projection certifies.

Deep generative models now downscale and forecast weather and climate fields. CorrDiff~\citep{mardani2023residual} uses diffusion to downscale a 25\,km reanalysis to 2\,km wind, temperature, and a radar-reflectivity surrogate for rain over Taiwan. GenCast~\citep{price2023gencast} uses diffusion for global ensemble forecasts at 0.25\textdegree, and data-driven forecasters such as GraphCast~\citep{lam2023graphcast} have changed medium-range prediction. Rectified flow~\citep{liu2022flow,lipman2022flow} straightens the path from noise to data and needs far fewer network evaluations per sample. A super-resolution network taken unchanged from natural images imposes no physics, and a pixel-wise or perceptual objective does not guarantee that the output respects the atmospheric physics of its coarse input. Generated wind can diverge, precipitation can go negative, and the domain mass can drift from the coarse input.

Physics can enter a learned model as a soft penalty~\citep{raissi2019physics} or as part of its architecture~\citep{kochkov2021machine,kohler2020equivariant}. For downscaling, constraint layers can enforce conservation exactly at the output of a network~\citep{harder2023hard}, and recent work adds soft per-cell conservation penalties for precipitation and humidity to an adaptive flow-matching downscaler~\citep{debeire2026physics}. Hard constraints at sampling time already exist for pretrained flow-matching models of PDE systems~\citep{cheng2025gradient,utkarsh2025physics} and for a consistency-model downscaler~\citep{bochow2026physics}. \citet{hess2022physically} add one global rescaling of precipitation to a trained generative adversarial network. \pcrf applies the same domain-level rescaling, after a clamp at zero, to the precipitation samples of a rectified-flow downscaler. The training penalties are meant to shape the generated field, and the inference-time projection makes its non-negativity and domain mass balance exact. The evidence in this paper supports the projection. It does not show that the penalties improve on an unconstrained rectified-flow baseline on reanalysis data, and we do not claim that they do.

\noindent\textbf{Contributions:} We make three contributions.
{\setlength{\topsep}{0pt}\setlength{\partopsep}{0pt}\setlength{\itemsep}{0pt}\setlength{\parsep}{0pt}\setlength{\parskip}{0pt}
\begin{itemize}
\item We propose Physics-Constrained Rectified Flow (\pcrf), a conditional rectified-flow downscaler that pairs three differentiable training penalties (on wind divergence, negative precipitation, and mass drift from the coarse total) with an inference projection that makes non-negativity and mass balance exact.
\item Next to SSIM, PSNR, and CRPS, we report three physics-validity metrics, the Non-negativity Violation Rate (NVR, the negative-pixel count of \citet{harder2023hard} as a fraction of cells), the Mass-Conservation Error (MCE, the domain-level counterpart of their per-cell mass violation, as a relative error of the domain mean), and the Divergence Error (DE, whose ERA5 values here measure a deformation-like term, as \cref{sec:experiments} explains), and we argue that NVR and MCE should run as a gate that needs no reference.
\item We evaluate the projection and the penalties separately on one synthetic field and on ERA5, compare \pcrf with bicubic interpolation and four reimplemented baselines on one year of ERA5 and on three years (2019--2021), and report that the projection certifies non-negativity and mass balance wherever it is applied, whereas the penalties give no perceptual gain that we can separate from training budget or run-to-run variation. We report that negative result.
\end{itemize}}

\section{Background and method}
\label{sec:background}
Conditional generative downscaling treats the fine field as a sample from a distribution conditioned on the coarse field. Denoising diffusion models~\citep{ho2020denoising,song2021score} turn a clean field into Gaussian noise through a forward chain and learn the reverse chain conditioned on a context $\bc$. Each sample then costs hundreds to thousands of network evaluations. Rectified flow~\citep{liu2022flow,lipman2022flow} replaces the stochastic reverse chain with a deterministic transport along straight paths from noise to data. We adopt it for its sampling efficiency at downscaling resolutions.

\textbf{Objective:} Rectified flow learns a velocity field $\bv_\theta$ that moves samples from a Gaussian source $p_0$ to the data distribution $p_1$ along straight paths. Given paired samples $(\bx_0,\bx_1)$, the linear interpolation $\bx_t = (1-t)\bx_0 + t\bx_1$ gives the objective
\begin{equation}
\mathcal{L}_{\mathrm{RF}} = \Eb{t, \bx_0, \bx_1}{ \norm{ \bv_\theta(\bx_t, t, \bc) - (\bx_1 - \bx_0) }_2^2 },
\label{eq:rf}
\end{equation}
where $\bc$ is the conditioning context. Here and in the penalties below, the squared norm is averaged over grid cells and channels. At inference we integrate the ODE $d\bx_t/dt = \bv_\theta(\bx_t, t, \bc)$ from $t{=}0$ to $t{=}1$ with a forward Euler solver in 20 steps.

\textbf{Problem formulation:} Let $\bx_c \in \R^{H_c \times W_c \times C}$ be a coarse climate field (ERA5 precipitation and $u$/$v$ wind at 0.25\textdegree), and let $\bx_f \in \R^{H_f \times W_f \times C}$ be the fine target on a finer grid over the same box. Our fine grid is a square $128 \times 128$ grid, about three times finer than the coarse grid in latitude and about two times finer in longitude. We train $\bv_\theta(\bx_t, t, \bc)$ conditioned on $\bc = \bx_c^{\uparrow}$, the coarse field bicubically upsampled to the fine grid. Four extra channels, reserved for satellite land-surface context, are zero in every reported run.

\textbf{Physics penalties:} We add three differentiable penalties to~\cref{eq:rf}. Each uses the one-step estimate $\tilde\bx_f = \bx_t + (1-t)\,\bv_\theta(\bx_t, t, \bc)$. Wind $(\tilde u, \tilde v)$ is penalized by its squared divergence,
\begin{equation}
\mathcal{L}_{\mathrm{div}} = \Ea{ \norm{ \tfrac{\partial \tilde u}{\partial x} + \tfrac{\partial \tilde v}{\partial y} }_2^2 },
\label{eq:div}
\end{equation}
with derivatives taken by central finite differences on the grid. The penalty pulls the wind toward the non-divergent approximation and acts on the total divergence, including any divergence already in the coarse wind. On ERA5 the same operator computed a deformation-like term, for the reason given in \cref{sec:experiments}. Because precipitation cannot be negative, we penalize $\mathcal{L}_{\mathrm{nn}} = \Ea{\norm{\min(\tilde p, 0)}_2^2}$. The generator should neither add nor remove precipitation at the domain scale, so the mass penalty is the squared gap of the two means,
\begin{equation}
\mathcal{L}_{\mathrm{mass}} = \Ea{ \left| \tfrac{1}{H_f W_f}\textstyle\sum_{i,j} \tilde p_{ij} - \tfrac{1}{H_c W_c}\textstyle\sum_{i',j'} p^c_{i'j'} \right|^2 }.
\label{eq:mass}
\end{equation}
The full loss adds the penalties to the flow objective, $\mathcal{L} = \mathcal{L}_{\mathrm{RF}} + \lambda_1 \mathcal{L}_{\mathrm{div}} + \lambda_2 \mathcal{L}_{\mathrm{nn}} + \lambda_3 \mathcal{L}_{\mathrm{mass}}$, with each $\lambda_i$ fixed (\cref{sec:experiments}).

\textbf{Why these constraints:} The three penalties describe physics that a downscaled field should respect before it is used downstream. The non-divergent approximation holds for synoptic-scale flow aloft. It is weaker for 10\,m wind, where friction, land-sea contrast, and orography make low-level divergence physically real, so the divergence penalty is a soft prior with no guarantee. Precipitation is an accumulation and cannot be negative. Domain mass balance ties the fine domain mean to the coarse domain mean, a single number that constrains the regional water budget but no individual basin. A field can violate these constraints and still score well on perceptual metrics, which is why we measure validity directly.

\textbf{Inference-time projection:} The penalties alone do not guarantee a valid field. After integrating the sampling ODE, we apply a projection $\proj$ to the precipitation channel $\hat p$ of the sampled field $\hat\bx_f$, and the wind channels pass through unchanged. The projection clamps precipitation at zero and then rescales it by one factor, which sets the domain mean equal to the coarse mean up to float32 rounding,
\begin{equation}
\begin{aligned}
\proj(\hat p)_{ij} &= s\,\hat p^{+}_{ij}, \qquad \hat p^{+}_{ij} = \max(\hat p_{ij},\, 0), \\
s &= \frac{m_c}{\max(m^{+},\, \varepsilon)}.
\end{aligned}
\label{eq:proj}
\end{equation}
Here $m_c$ is the domain mean of the coarse precipitation, $m^{+}$ is the domain mean of the clamped field $\hat p^{+}$, and $\varepsilon = 10^{-8}$ guards the division when the clamped field is dry everywhere. \Cref{eq:proj} makes NVR exactly zero for any model. It also makes MCE zero whenever the clamped field has at least one positive cell, up to floating-point rounding far below the precision we report. The zeros we report show that an all-dry clamped field under a wet coarse mean never occurred. Below, ``zero NVR and MCE'' means exactly zero NVR and MCE zero up to float32 rounding.

\textbf{What the projection changes:} The projection consists of two operations, and neither can create structure. The clamp sets every negative precipitation cell to zero, which changes only those cells and raises the domain mean. The whole channel is then multiplied by one factor per sample, the ratio of the coarse mean to the mean of the clamped field. A single global factor keeps every ratio between cells and multiplies every spatial gradient by the same amount, so the rescaling changes the amplitude and leaves the pattern in place. The clamp changes that pattern only in cells that were negative.

The size of the local change follows from \cref{eq:proj}. A negative cell moves to zero, which is a change equal to its own magnitude, and a dry cell stays dry. Every wet cell changes by the same relative amount, $|s-1|$. Consider a time step with $m_c > 0$, and let $\bar p$ be the domain mean of the sampled precipitation before the projection. Let $e = (\bar p - m_c)/m_c$ be its relative mass error, and let $n$ be the mean magnitude of its negative values, taken over all cells and divided by $m_c$. The clamped mean is then $m^{+} = m_c\,(1 + e + n)$, so
\begin{equation}
s = \frac{1}{1 + e + n}.
\label{eq:scalar}
\end{equation}
The relative change in every wet cell depends on two properties of the unprojected sample, its mass error and its share of negative mass. A sample with neither is left unchanged. For a field scored without the projection, MCE is the test-set average of $|e|$.

On our data, the reference that a model learns from is zero over the sea and accumulates over a longer window than the coarse input (\cref{sec:experiments}). A trained model therefore need not match the coarse mean, and the factor rescales every cell to close that gap. The unprojected baselines learn the same reference, and their MCE ranges from 6.5 to 13.2 in \cref{tab:era5}, so our own factor may be far from one. Where a model keeps the sea dry, as the reference does, the rescaled land cells also receive the share of the coarse mean that fell over the sea. Because we did not save the field before projection, we recorded neither the fraction of cells that the clamp changes nor the distance of the factor from one, which \cref{eq:scalar} ties to $e$ and $n$. One sampling pass with the projection switched off would give both, but we did not run it. For scale, the baselines that we score without the projection reach NVR values from $0.053$ to $0.826$ in \cref{tab:era5}. None of them uses our non-negativity penalty, so those rates do not bound what our own sampler produces.

\textbf{Why the wind is not projected:} Non-negativity is a pointwise constraint and domain mass is a single scalar, so each has a closed-form projection that costs one pass over the field. A divergence-free projection has no such closed form. It requires solving a Poisson equation for a velocity potential and subtracting its gradient, which is a global operation over the whole grid. Our setup also has no fine wind target. The wind channels hold the upsampled coarse wind, and projecting them would move the field away from the only reference that the perceptual scores use. Finally, zero horizontal divergence is an approximation and not an exact law, because low-level convergence is physically real and mass continuity turns it into ascent. Forcing it to zero would remove signal that belongs in the field. On ERA5 the divergence penalty acted on the deformation-like term of \cref{sec:experiments}, and we cannot separate its measured effect from run-to-run variation (\cref{sec:certify}). Because the wind is left unprojected, we report DE as a diagnostic and treat only non-negativity and mass balance as certificates.

\begin{figure*}[t]
\centering
\includegraphics[width=\textwidth]{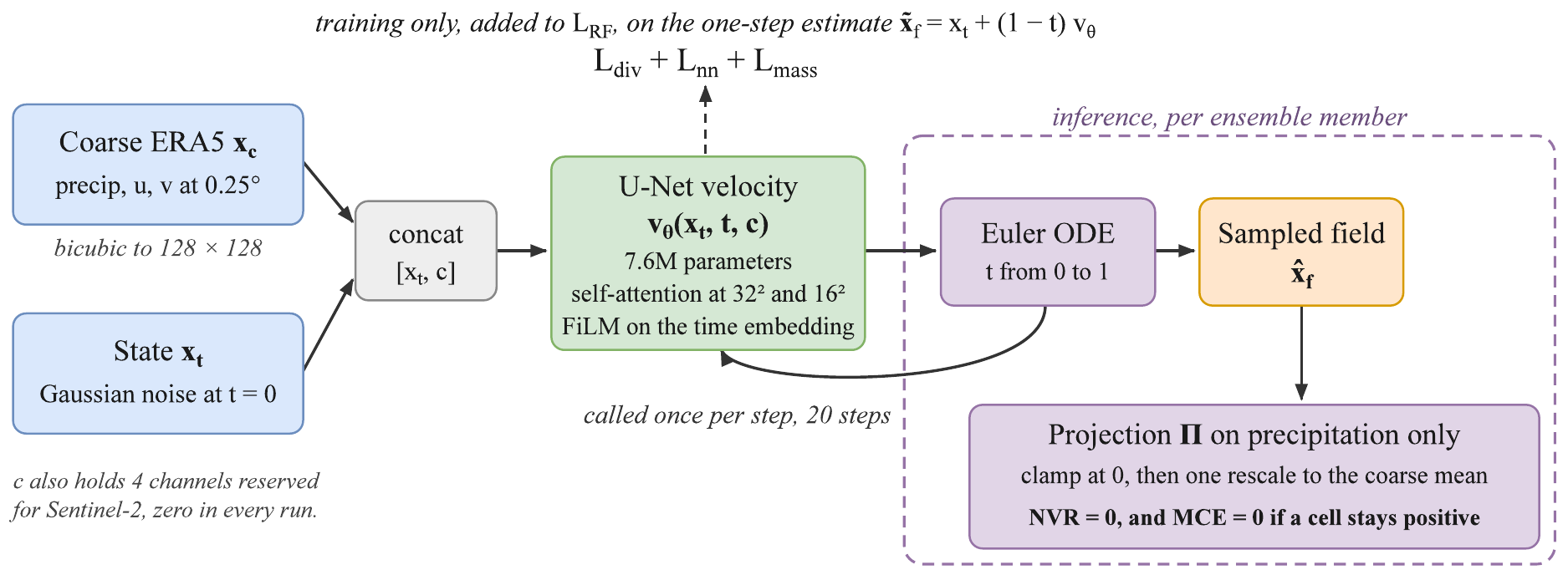}
\Description{A block diagram that reads left to right. The coarse ERA5 field is bicubically upsampled and concatenated with the noisy state. Both enter a U-Net velocity network with a time embedding. During training, the flow loss and three physics penalties act on a one-step estimate of the clean field. At inference, a 20-step Euler solver produces a sampled field, and a projection step clamps precipitation at zero and rescales it to the coarse mean.}
\caption{The trained \pcrf pipeline. The coarse ERA5 field $\bx_c$ is upsampled and concatenated with $\bx_t$. The U-Net $\bv_\theta$ uses self-attention only at its two coarsest resolutions and FiLM only for time. Training adds three penalties on the one-step estimate $\tilde\bx_f$. Inference takes 20 Euler steps to $\hat\bx_f$, and $\proj$ then makes precipitation non-negative and mass-balanced.}
\label{fig:arch}
\end{figure*}

\textbf{Architecture:} The velocity network $\bv_\theta$ is a U-Net~\citep{ronneberger2015u} with residual blocks, skip connections, and three resolution levels (128, 64, and 32 cells on a side) above a $16 \times 16$ bottleneck. A sinusoidal time embedding enters every residual block, and a FiLM layer~\citep{perez2018film} also uses it to modulate the first feature map. Multi-head self-attention runs at the $32 \times 32$ level and in the bottleneck. The first convolution takes $\bx_t$ concatenated with the conditioning image $\bc$. The code reserves the four extra conditioning channels for Sentinel-2 Level-2A land-surface context~\citep{drusch2012sentinel}. Every run we report fills them with zeros and trains no land-surface encoder, so no land-surface information reaches the model and we report no Sentinel-2 results. The trained model has 7.6M parameters, small by design. \Cref{fig:arch} summarizes the pipeline.

\section{Experiments}
\label{sec:experiments}
We ask whether physics constraints reduce violations of physical laws without lowering perceptual quality. Where the design allows, we separate the effect of the training penalties from that of the inference projection. We report SSIM, PSNR, and probabilistic CRPS as perceptual scores, computed over the three channels together, and three physics-validity metrics. A sample is one time step over the whole box. For the sampled models (the two flow models, DDPM, and CorrDiff), SSIM, PSNR, and the three validity metrics are computed on the mean of 8 ensemble members, and CRPS on the members themselves. The deterministic baselines are scored on their single output. Because averaging smooths a field, these scores are not strictly comparable across the two groups. The SSIM and PSNR data range comes from the reference in each consecutive test batch of up to 16 samples. Every score is a mean of per-batch means, so a short final batch weighs as much as a full batch.

Divergence Error (DE) is the mean $|\partial \hat u/\partial x + \partial \hat v/\partial y|$ over the test set, taken by central differences in grid units on the standardized wind channels, which makes it dimensionless. Non-negativity Violation Rate (NVR) is the fraction of precipitation cells with $\hat p < 0$, counted with no tolerance. Precipitation enters the model divided by the standard deviation of the coarse training precipitation, with no mean removed, so zero stays zero and NVR counts truly negative values. The perceptual scores are computed in the same scaled units, and CRPS therefore has no physical unit. Mass-Conservation Error (MCE) is the relative error between the domain-mean fine precipitation and the coarse mean. It is computed per sample and averaged over the test set, so a nearly dry time step counts for more.

\textbf{Data:} The coarse input is total precipitation and 10\,m $u$/$v$ wind from ERA5, the ECMWF fifth-generation reanalysis, at 0.25\textdegree~\citep{hersbach2020era5}. We use a western European box (40 to 50\textdegree N, 10\textdegree W to 5\textdegree E) at 00, 06, 12, and 18 UTC for 2020. The fine precipitation target is ERA5-Land total precipitation at 0.1\textdegree\ over the same box and hours. We resample it to the square $128 \times 128$ grid, about 0.08\textdegree\ by 0.12\textdegree\ per cell, and set the sea cells of that land-only product to zero. We use no fine wind target, and the fine wind channels hold the bicubically upsampled coarse wind. This wind reference has a nonzero DE of its own, which is the DE of the bicubic row in \cref{tab:era5} and \cref{tab:era5_3yr}. Because bicubic reproduces the wind reference exactly, its gaps to the learned models in SSIM, PSNR, and CRPS come from precipitation and from any wind error of those models.

The precipitation pair limits what the scores below can show. The fine field is not independent of the coarse one. ERA5-Land replays the land component at 0.1\textdegree\ under ERA5 forcing~\citep{munozsabater2021era5land}. Total precipitation is one of the forcing fields that it inherits and is not a land-model output, and ECMWF states that such fields are purely interpolated onto the finer grid~\citep{ecmwf2025era5land}. Our fine reference therefore has no precipitation structure that ERA5 does not already have. The two products also follow different accumulation conventions~\citep{ecmwf2025era5land}. An ERA5 value covers the hour ending at the validity time~\citep{ecmwf2026era5}. An ERA5-Land value accumulates from 00 UTC, so at 06, 12, and 18 UTC it covers 6, 12, or 18 hours, and at 00 UTC it covers the previous 24 hours. The coarse and fine precipitation that we compare never cover the same interval, and we do not de-accumulate either field. The mass penalty, the projection, and MCE all use a coarse mean taken over land and sea alike. Over the sea the zero-filled reference has less water than the coarse total, and over land its longer window can give it more. A zero MCE therefore certifies agreement with the coarse input alone.

\textbf{Divergence on the ERA5 grid:} Our divergence operator has a sign error on ERA5. The ERA5 and ERA5-Land arrays keep the latitude order of the source files, so row 0 is the northern edge of the box, and a central difference taken down the row index is the negative of the northward derivative. The operator in \cref{eq:div} and in DE therefore computes $\partial u/\partial x - \partial v/\partial y$ on ERA5, which is a stretching-deformation term and not the divergence. The two wind channels are also standardized separately, and the $10^\circ \times 15^\circ$ box is resampled to a square grid. Even with the correct sign, the two derivative terms would have unequal weights, set by the two standard deviations and the cell aspect, and their sum would not be proportional to the physical divergence. Every ERA5 DE value and the ERA5 divergence penalty refer to that quantity, which we call the deformation-like term. The synthetic check builds its fields in grid coordinates and is not affected.

\textbf{Training and baselines:} We split the year chronologically 80/10/10 into training, validation, and test sets. The test set is the final tenth of the year, from 25 November to 31 December 2020. No reported run uses the validation tenth, because no run has a validation loop and every model is scored from its final-epoch checkpoint. The training split has 1171 samples, which gives 36 optimizer steps per epoch at batch 32. The velocity network is the 7.6M-parameter U-Net of \cref{sec:background}. We train it with AdamW for 300 epochs at batch size 32, learning rate $2{\times}10^{-4}$, and physics weights $\lambda_1{=}\lambda_2{=}\lambda_3{=}1.0$ on one NVIDIA A100-40GB. We sample the rectified-flow models with 20 forward Euler steps and 8 ensemble members. The DDPM and CorrDiff reimplementations use 100 stochastic DDIM steps ($\eta = 1$) and 8 members. The deterministic baselines give a single member, for which CRPS reduces to the mean absolute error.

Bicubic interpolation needs no training. SRCNN~\citep{dong2015image}, SRGAN~\citep{ledig2017photo}, a DDPM downscaler~\citep{ho2020denoising}, and CorrDiff~\citep{mardani2023residual} are our reimplementations and not the original authors' released models. We train them on the same split and score them on the same held-out test set without the projection. The training budgets are not equal. \pcrf and the four learned baselines train for 300 epochs. SRCNN, SRGAN, and DDPM use batch 32, and CorrDiff uses batch 16 to fit its two U-Nets in 40\,GB, so CorrDiff takes about twice as many optimizer steps. RF-base is the rectified-flow baseline trained without the penalties. Its row is the no-physics arm of the ablation run (\cref{tab:ablation}), trained for 200 epochs at batch 16. It sees fewer samples than \pcrf but takes more optimizer steps (14{,}600 against 10{,}800). Every trained ERA5 model in this paper comes from a single training run, and neither its training nor its sampling uses a fixed seed.

\textbf{Synthetic mechanism check:} Before turning to real reanalysis, we test the constraints on one synthetic $128 \times 128$ field whose ground truth is physically exact by design (divergence-free wind from a stream function, non-negative precipitation from Gaussian blobs, a known coarse mass). \Cref{tab:synthetic} compares a corrupted baseline field with the same field after the \pcrf mechanism is applied. The baseline is the truth plus spatially correlated noise on all three channels, and its precipitation also has a constant negative bias of 0.18 times the standard deviation of the true precipitation. The \pcrf arm runs 600 gradient steps on the three penalties, weighted 2, 4, and 6 against a fidelity term of weight 1 that keeps the field near the baseline, and then applies the projection of \cref{eq:proj}. No network is trained in this check. Relative to the baseline, the penalties lower DE by about 30\% ($0.0081\to0.0057$). The projection cannot explain that change, because it never touches the wind. It removes all non-negativity violations (NVR $0.236\to0$) and drives MCE to zero. Precipitation fidelity improves as well (correlation $0.956\to0.966$, SSIM $0.518\to0.682$). The check cannot say how much of that gain the penalties produced, because the \pcrf arm applies the penalties and the projection together. The projection clamps the field and rescales it to the true coarse mean, which removes the net shift in the domain mean left by the bias and the clamp. The synthetic results confirm the constraints on one field, and we keep them apart from the ERA5 results.

\begin{table*}[t]
\caption{One synthetic $128 \times 128$ field, fixed seed, no trained network. These rows are not ERA5 results. RMSE, Corr, and SSIM score precipitation only, and the $7 \times 7$ SSIM window is not comparable with \cref{tab:era5,tab:ablation,tab:era5_3yr}. Bold is better ($\uparrow$ higher, $\downarrow$ lower).}
\label{tab:synthetic}
\centering
\begin{tabular}{lcccccc}
\toprule
\textbf{Method} & DE$\downarrow$ & NVR$\downarrow$ & MCE$\downarrow$ & RMSE$\downarrow$ & Corr$\uparrow$ & SSIM$\uparrow$ \\
\midrule
Ground truth & 0.0000 & 0.000 & 0.0000 & 0.0000 & 1.000 & 1.000 \\
Corrupted field (no physics) & 0.0081 & 0.236 & 0.1838 & 0.1397 & 0.956 & 0.518 \\
\textbf{\pcrf mechanism (physics)} & \textbf{0.0057} & \textbf{0.000} & \textbf{0.0000} & \textbf{0.1116} & \textbf{0.966} & \textbf{0.682} \\
\bottomrule
\end{tabular}
\end{table*}

\textbf{Real ERA5 downscaling:} \Cref{tab:era5} compares \pcrf and RF-base with bicubic interpolation and our reimplemented baselines on the single-year task. No model wins every column, since CorrDiff has the best PSNR (28.38) and CRPS (1.092) while \pcrf has the best SSIM (0.914). RF-base and \pcrf both use the inference projection, so they are the only two models with zero NVR and MCE. Every learned baseline without the projection leaves between 5\% and 83\% of its precipitation cells negative (NVR 0.053 to 0.826, on the ensemble mean for DDPM and CorrDiff) and has an MCE between 6.5 and 13.2. We did not score the reference for MCE. It departs from the coarse mean (Data paragraph), so a model that reproduced it exactly would not score zero either, and these values mix model error with that offset. Bicubic is the one exception on mass (MCE 0.0003), because its kernel nearly preserves the coarse mean, although its ringing still makes 10\% of cells negative. A negative accumulation is physically impossible, and a hydrological model that received such a field would have to clip it first.

On the 2020 test set, \pcrf also scores above RF-base (SSIM $0.912\to0.914$, PSNR $26.39\to26.59$, CRPS $1.188\to1.149$), but the two rows had different training budgets. In the matched-budget ablation of \cref{tab:ablation} the SSIM gap closes (0.912 for full \pcrf against 0.913 for RF-base), while full \pcrf keeps a smaller PSNR and CRPS lead (26.54 against 26.40, and 1.161 against 1.186). Every penalty arm leads RF-base on those two scores on 2020. With one run per arm we cannot tell that lead apart from run-to-run variation, and on 2019--2021 no consistent lead remains (\cref{tab:ablation}). The tables here and in \cref{tab:era5_3yr} show which models stay physically valid, not which one downscales better, given the reference limits in the Data paragraph.

\begin{table}[t]
\caption{ERA5 2020, precipitation and two wind channels. The wind target is the upsampled coarse wind. Learned baselines are our reimplementations (300 epochs, no projection). RF-base used 200 epochs at batch 16, and \pcrf used 300 epochs at batch 32. Only the projected models have zero NVR and MCE. Bicubic DE is the wind reference. DE is the deformation-like term of \cref{sec:experiments}, so it has no arrow or bold. Bold is the best other value ($\uparrow$ higher, $\downarrow$ lower).}
\label{tab:era5}
\centering
\small
\begin{tabular}{lcccccc}
\toprule
\textbf{Method} & SSIM$\uparrow$ & PSNR$\uparrow$ & CRPS$\downarrow$ & DE & NVR$\downarrow$ & MCE$\downarrow$ \\
\midrule
Bicubic & 0.885 & 26.19 & 1.356 & 0.0648 & 0.101 & 0.0003 \\
SRCNN & 0.833 & 27.74 & 1.461 & 0.0738 & 0.053 & 11.1365 \\
SRGAN & 0.855 & 27.03 & 1.506 & 0.0700 & 0.177 & 13.2239 \\
DDPM & 0.656 & 26.18 & 1.609 & 0.0715 & 0.826 & 7.5011 \\
RF-base & 0.912 & 26.39 & 1.188 & 0.0643 & \textbf{0.000} & \textbf{0.0000} \\
CorrDiff & 0.864 & \textbf{28.38} & \textbf{1.092} & 0.0679 & 0.451 & 6.4835 \\
PC-RF & \textbf{0.914} & 26.59 & 1.149 & 0.0644 & \textbf{0.000} & \textbf{0.0000} \\
\bottomrule
\end{tabular}
\end{table}

\textbf{Three-year ERA5 downscaling:} To test the penalties at a matched budget and with more data, we retrained every learned model on ERA5 for 2019--2021, about three times the training data. Every learned model trains for 100 epochs, against 300 and 200 on 2020. The two rectified-flow models and CorrDiff train at batch 16, and SRCNN, SRGAN, and DDPM train at batch 32. The chronological split again holds out the final tenth of the record, from 13 September to 31 December 2021, and \cref{tab:era5_3yr} reports the test metrics. \pcrf has a DE of 0.0563, against 0.0567 for bicubic and 0.0570 for RF-base. That gap is smaller than the spread of the three-year ablation arms (0.0558 to 0.0586, \cref{tab:ablation}). RF-base uses the same projection and trains without penalties, and it matches \pcrf on every perceptual score (SSIM 0.942 vs.\ 0.939, PSNR 31.79 vs.\ 31.81, CRPS 0.772 vs.\ 0.771). Both models have zero NVR and MCE. CorrDiff again has the best PSNR (32.02) and CRPS (0.761), but its MCE grows to 32.4, and the MCE of DDPM reaches 170.2. As on 2020, these large values mix model error with the offset of the reference. At a matched budget, we cannot tell the perceptual effect of any penalty from run-to-run variation, and the projection supplies the validity guarantee.

\textbf{Ablation:} \Cref{tab:ablation} removes one physics penalty at a time. The one-year full arm is a separate run from the 300-epoch \pcrf of \cref{tab:era5}. NVR and MCE are zero in every row, including RF-base, because the projection enforces them after sampling. On 2020, removing the divergence penalty changes DE the most, from 0.0634 for the full model to 0.0649, a small change in the expected direction. RF-base trains with no penalty at all, and its DE of 0.0643 lies between the two. Removing the non-negativity or mass penalty moves SSIM by about 0.003 or less (0.912 to 0.915 across all arms). The three-year block adds an arm that drops the four zero-filled satellite channels from the input. Those channels hold no information, so the arm sees what the full model sees and differs from it only in input width and in the random draws of an unseeded run. Its DE is 0.0586 against 0.0563, and its CRPS is 0.782 against 0.771. Because that DE gap is larger than the change from removing the divergence penalty at either scale, we cannot separate the DE changes in this table from variation between runs. Removing the mass penalty on 2019--2021 gives the lowest DE of any arm (0.0558) and an SSIM of 0.942, level with RF-base at the precision we report.

\begin{table*}[t]
\caption{Ablations at 200 epochs, batch 16 (2020) and 100 epochs, batch 16 (2019--2021). Each named row drops one penalty. The satellite arm (three-year only) drops four zero-filled inputs. The 2020 RF-base checkpoint is that row of \cref{tab:era5}, and the three-year full and RF-base checkpoints are those rows of \cref{tab:era5_3yr}, with a new 8-member sample. Zero NVR and MCE are omitted. DE is unmarked, as in \cref{tab:era5}. Bold marks the best SSIM, PSNR, and CRPS per block.}
\label{tab:ablation}
\centering
\begin{tabular}{lcccc@{\hspace{10pt}}cccc}
\toprule
 & \multicolumn{4}{c}{2020 (one year)} & \multicolumn{4}{c}{2019--2021 (three years)} \\
\cmidrule(lr){2-5}\cmidrule(lr){6-9}
\textbf{Configuration} & SSIM$\uparrow$ & PSNR$\uparrow$ & CRPS$\downarrow$ & DE & SSIM$\uparrow$ & PSNR$\uparrow$ & CRPS$\downarrow$ & DE \\
\midrule
Full \pcrf                  & 0.912 & 26.54 & 1.161 & 0.0634 & 0.939 & 31.80 & \textbf{0.771} & 0.0563 \\
$-$ divergence penalty      & 0.912 & \textbf{26.58} & \textbf{1.146} & 0.0649 & 0.939 & 31.78 & 0.773 & 0.0572 \\
$-$ non-negativity penalty  & 0.913 & 26.55 & 1.160 & 0.0644 & 0.940 & 31.86 & 0.771 & 0.0570 \\
$-$ mass penalty            & \textbf{0.915} & 26.50 & 1.166 & 0.0639 & 0.942 & 31.80 & 0.783 & 0.0558 \\
$-$ satellite channels      & n/a & n/a & n/a & n/a & 0.939 & \textbf{31.89} & 0.782 & 0.0586 \\
RF-base (no physics)        & 0.913 & 26.40 & 1.186 & 0.0643 & \textbf{0.942} & 31.79 & 0.772 & 0.0570 \\
\bottomrule
\end{tabular}
\end{table*}

\begin{table}[t]
\caption{ERA5 for 2019--2021, scored as in \cref{tab:era5} on the final tenth, with learned models at 100 epochs. Reimplemented baselines have no projection. Only RF-base and \pcrf have zero NVR and MCE. DE is unmarked, as in \cref{tab:era5}. Bold marks the best value ($\uparrow$ higher, $\downarrow$ lower).}
\label{tab:era5_3yr}
\centering
\small
\begin{tabular}{lcccccc}
\toprule
\textbf{Method} & SSIM$\uparrow$ & PSNR$\uparrow$ & CRPS$\downarrow$ & DE & NVR$\downarrow$ & MCE$\downarrow$ \\
\midrule
Bicubic & 0.927 & 31.48 & 0.873 & 0.0567 & 0.119 & 0.0006 \\
SRCNN & 0.832 & 30.43 & 1.057 & 0.0618 & 0.032 & 47.8114 \\
SRGAN & 0.869 & 31.49 & 0.888 & 0.0601 & 0.324 & 13.7981 \\
DDPM & 0.676 & 28.61 & 1.173 & 0.0648 & 0.054 & 170.1641 \\
RF-base & \textbf{0.942} & 31.79 & 0.772 & 0.0570 & \textbf{0.000} & \textbf{0.0000} \\
CorrDiff & 0.820 & \textbf{32.02} & \textbf{0.761} & 0.0598 & 0.108 & 32.3586 \\
PC-RF & 0.939 & 31.81 & 0.771 & 0.0563 & \textbf{0.000} & \textbf{0.0000} \\
\bottomrule
\end{tabular}
\end{table}

\section{What the validity metrics can certify}
\label{sec:certify}
Two of the three validity metrics are certificates. After the projection of \cref{eq:proj}, NVR and MCE are zero whatever objective the model was trained with, so a zero in those columns describes the inference step and says nothing about the learned weights. A certificate also cannot rank two models that share it, as \cref{tab:era5_3yr} shows. \pcrf and RF-base pass through the same projection, both report NVR 0.000 and MCE 0.0000, and their perceptual scores differ by about 0.003 in SSIM, 0.02\,dB in PSNR, and 0.001 in CRPS.

DE, the third metric, is a diagnostic. The projection clamps and rescales precipitation and never touches the wind. Nothing in the method forces the divergence error toward zero. On the synthetic field, where DE measures the true divergence, the penalty lowers it by about 30\% (0.0081 to 0.0057). On ERA5 2020 the arm without the divergence penalty is about 2\% above the full model (0.0649 against 0.0634), while two three-year arms that differ only in four zero-filled input channels are about 4\% apart (0.0586 against 0.0563), so we cannot attribute the ERA5 change to the penalty. The ERA5 values also measure the deformation-like term. The meaning of a DE value depends on the reference as well. On the synthetic field the true wind is divergence-free, and any nonzero DE is spurious. On ERA5 the fine wind reference is the upsampled coarse wind, whose own DE is the bicubic row's 0.0648 on 2020 and 0.0567 on 2019--2021. The flow models of \cref{tab:era5} and \cref{tab:era5_3yr} are within 1\% of that reference level, and some are slightly below it. Averaging 8 members smooths the wind and can place DE below that level.

We propose NVR and MCE as a gate, a check to run before any other score is read. A field passes when it has zero NVR and MCE. A pass says that the field is non-negative and matches the coarse domain mean. Validity is necessary but not sufficient, because non-negativity fixes a sign at each cell and mass balance fixes one scalar per domain, and nearly every other degree of freedom stays open. On our data pair the mass relation that the projection enforces is also not the physical one, so an MCE of zero here certifies agreement with the coarse ERA5 total and nothing more. SSIM, PSNR, and CRPS then rank the fields that pass the gate. Every unprojected model in \cref{tab:era5} and \cref{tab:era5_3yr} fails it, bicubic included, because we scored them without the projection that any of them could apply. The fine reference would not pass the MCE check either, for the reasons given in \cref{sec:experiments}.

The checks are cheap, because NVR, MCE, and DE need only the coarse input and the generated field and take one pass with no simulation. They should run first when verification time is limited, and they still work when no reference exists yet. An agent for spatial question answering shows a related pattern, computing in code what a language model would estimate and withholding the answer when an evidence check fails~\citep{sharma2026spatial}.

\textbf{Value to practice:} The perceptual differences between the two flow models are small, and we do not offer them as a reason to use the method. Two parts of the work can be used as they are. The projection of \cref{eq:proj} costs a clamp and one division per field and needs no retraining. It can follow any generative downscaler whose output feeds a hydrological or impact model, provided that the coarse and fine fields cover the same area and the same accumulation window. The gate gives a pass-or-fail answer for a field that has no reference yet. Our data cannot settle what the penalties add once the projection is in place, for the reasons given next.

\textbf{Limitations:} The most serious limitation is the data design. Our fine reference is interpolated from ERA5, the same reanalysis that supplies our coarse input, and it has no precipitation structure of its own. It also differs from that input in its accumulation window and in its zero-filled sea cells (Data paragraph). The reference therefore does not satisfy the coarse-to-fine mass relation that our penalty and projection aim at, and we corrected neither mismatch. The null result of \cref{tab:era5_3yr} has two possible explanations that we cannot separate here. A soft conservation penalty may add nothing once the constraint is enforced at inference. The penalty may also fail to help because this pair of products does not satisfy the relation that it enforces. This paper does not test the mass penalty on an independent target over matched accumulation windows.

The other limitations concern scope and replication. We report a single western European domain, variable set, and target grid. We ran no evaluation conditioned on extremes, although these checks matter most on days of heavy rain and strong wind. Both test sets fall in autumn and early winter. We report no seasonal or regional breakdown and no test of generalization under distribution shift, because the chronological split makes every test set the final stretch of the record. Such a breakdown would need statistical care of its own. A spatial pattern that is statistically significant in one region can be absent in another~\citep{ghosh2022geographically}, and many simultaneous regional tests inflate false discoveries unless they are corrected~\citep{ghosh2023false}, for example with the Benjamini--Hochberg procedure~\citep{ghosh2024taxonomy}. Every arm is one training run with no confidence intervals, so SSIM differences of a few thousandths cannot be separated from run-to-run variation. Training budgets are not matched across models either (\cref{sec:experiments}), and the comparison across data sizes changes the training budget and the test period along with the data. The baselines are our reimplementations and were scored without the projection. Whether a projected CorrDiff keeps its PSNR and CRPS lead is untested. On ERA5 the divergence penalty and DE act on the deformation-like term, and these runs do not measure the physical divergence of the generated wind. No Sentinel-2 input reached any model.

\section{Related work}
\label{sec:related}
Classical statistical downscaling~\citep{maraun2018statistical} captures spatial structure but assumes stationarity and oversmooths fine-scale detail. Convolutional networks~\citep{dong2015image} and then adversarial networks~\citep{goodfellow2014generative,ledig2017photo} drove single-image super-resolution, and both were adapted to the geosciences. Stacked super-resolution networks downscale climate projections~\citep{vandal2017deepsd}, and adversarial networks recover fine-scale wind and solar structure~\citep{stengel2020adversarial} and downscale time-evolving fields with calibrated spread~\citep{leinonen2020stochastic}. Normalizing flows~\citep{rezende2015variational,dinh2016density,kingma2018glow} give exact likelihoods and have been applied to unsupervised climate downscaling~\citep{groenke2020climalign}. Denoising diffusion~\citep{ho2020denoising,song2021score} now underlies regional downscalers such as CorrDiff~\citep{mardani2023residual} and global ensemble forecasters such as GenCast~\citep{price2023gencast}. \pcrf builds on rectified flow~\citep{liu2022flow,lipman2022flow}, which samples in far fewer steps on image benchmarks.

Some methods make a constraint hard. \citet{harder2023hard} add constraint layers to deterministic and generative downscalers that make the fine-grid mean inside every coarse cell equal the coarse value exactly, and their softmax layer also keeps every output non-negative. Most of their layers improve fidelity over that same network. Their multiplicative layer can apply any function to the predictions before it rescales them, and a positive function guarantees a positive output. They also note that the constraint window can be widened until it covers the whole domain. \citet{hess2022physically} use one global constraint of that kind, a single factor that rescales the output of a trained generative adversarial network so that its global precipitation total matches its Earth system model input. \citet{bochow2026physics} enforce conservation on the coarse grid at inference in a consistency-model downscaler. Within flow matching, ECI sampling~\citep{cheng2025gradient} and PCFM~\citep{utkarsh2025physics} correct intermediate states during sampling so that pretrained models of PDE systems satisfy hard constraints exactly. \citet{beucler2021enforcing} compare loss-based and architecture-based constraints in neural emulators of convection and find that the architectural form holds conservation laws to machine precision without degrading skill. Outside climate, DC3~\citep{donti2021dc3} enforces equality constraints by completion and inequality constraints by unrolled gradient steps. In trajectory analysis, a hard physical bound stands in for missing observations. A space-time prism built from the maximum speed of a moving object bounds where it could have been during a gap in its record, which supports the search for possible rendezvous regions~\citep{sharma2020analyzing,sharma2022analyzing} and the detection of abnormal gaps~\citep{sharma2022abnormal,sharma2024physics}. Later work tightens the bound~\citep{sharma2022tighter} and pairs gap detection with satellite radar and optical imagery to find vessels that stop reporting their position~\citep{sharma2026darkvesselnet}. The projection in~\cref{eq:proj} applies the domain-level rescaling of \citeauthor{hess2022physically} to the clamped field. It is weaker than the per-cell layers of \citeauthor{harder2023hard}, and redistribution inside the domain stays free.

Most generative downscalers treat physical consistency as a diagnostic. Physics-informed learning~\citep{raissi2019physics} adds governing equations as soft penalties, and learned solvers and equivariant flows~\citep{kochkov2021machine,kohler2020equivariant} build physical structure into the architecture, but none of the three addresses conditional downscaling. Diffusion models have also taken in domain knowledge, such as a geostatistical prior in a sea-level downscaler~\citep{ghosh2024kriging}, kinematic constraints that flag trajectories that break the laws of motion~\citep{sharma2025physics}, and road-map attributes that raise the fidelity of generated vehicle trajectories~\citep{yang2025geolucid}. \citet{farhadloo2025physics} outline how physical knowledge can enter the data, the training, and the design of generative foundation models. In knowledge-guided machine learning, location-dependent parameters~\citep{sharma2025spatial} and site-level fine-tuning~\citep{zeng2025finetuning} raise the local accuracy of agroecosystem models, and a surrogate of the Daycent ecosystem model gains accuracy from explicit spatial autocorrelation and teleconnections~\citep{zhang2025surrogate}. Our single-domain study does not test that kind of spatial variability. The closest flow-matching relative of \pcrf is adaptive flow matching with soft per-cell conservation penalties~\citep{debeire2026physics}. \pcrf differs by making two of its constraints exact with a projection applied after sampling.

\section{Conclusion}
\label{sec:conclusion}
We introduced \pcrf, a rectified-flow downscaler whose three differentiable training penalties target wind divergence, negative precipitation, and mass drift from the coarse total, and whose projection at inference removes the last two exactly. On the held-out final tenth of 2020 and of 2019--2021, the two projected models are the only ones with an NVR of exactly zero and an MCE of zero up to float32 rounding, whether or not they were trained with the penalties. Our CorrDiff reimplementation has the best PSNR and CRPS on the 2020 test set but leaves 45\% of its ensemble-mean precipitation cells negative. On one synthetic field the penalties lower DE and the projection removes every NVR and MCE violation. On 2019--2021 at a matched budget, however, RF-base matches \pcrf on every perceptual score. Our reanalysis pair cannot settle what the penalties are worth, because its fine reference is interpolated from ERA5 and covers a different accumulation window from our coarse input. On ERA5 the divergence operator also took the northward derivative with the wrong sign, so those runs penalized and measured a deformation-like term. We propose NVR and MCE as a gate that needs no reference and runs before any perceptual score, with a correctly computed DE as a diagnostic. A field that fails the gate needs a correction before downstream use. The projection is one such correction when the coarse and fine fields cover the same area and the same accumulation window.

\textbf{Future work:} A regional reanalysis that produces its own precipitation, such as CERRA at 5.5\,km over Europe~\citep{ridal2024cerra}, would give a fine target that is not interpolated from ERA5, and its offline surface analysis blends that field with rain-gauge observations. Our first step is such a reference with matched accumulation windows. We will also correct the divergence operator so that it takes the northward derivative with the right sign, acts on the winds before standardization, and uses the physical spacing of the non-square cells. We will then penalize divergence relative to the upsampled coarse wind and retrain. After that we will score every baseline through the same projection, run every arm over multiple seeds at a matched budget, and extend the study to other regions, seasons, and extreme events. A model moved to a new region meets spatial variability that it was not trained on. Classifiers of spatial point patterns in oncology handle that problem with spatial domain adaptation across place types~\citep{farhadloo2024spatially,farhadloo2025spatially}, and a downscaler may need a similar step. The significance of a spatial pattern also depends on how its types are grouped and where it is measured. Super-colocation patterns and taxonomy-aware colocations change which groupings count once feature types are nested~\citep{an2025super,gupta2022taxonomy}. The same place dependence appears in public-health mobility, in a community study of {COVID-19} and a {SafeGraph} report on Minnesota travel~\citep{sharma2022covid,sharma2021safegraph}, and in multi-level bi-objective routing, where the preferred path changes with the objectives~\citep{yang2024routing}. A Sentinel-2 evaluation and a thermodynamic-consistency penalty are also planned.

\textbf{Broader impact:} Our target is scientific decision support. Negative precipitation or mass drift can mislead a hydrological model, and the projection reduces that risk. Climate model output also reaches policymakers and the public through browser-based analysis frameworks~\citep{sharma2018webglobe}, which is one more reason to check a field before it is served. \pcrf was trained on one domain and scored on two held-out test sets that both fall in autumn and early winter. It should not drive operational forecasting before broader evaluation across regions, seasons, and extremes.


\bibliographystyle{ACM-Reference-Format}
\bibliography{refs}

\end{document}